\documentclass[9pt,conference]{IEEEtran}
\usepackage[utf8]{inputenc} % allow utf-8 input
\usepackage[T1]{fontenc}    % use 8-bit T1 fonts
\usepackage{hyperref}       % hyperlinks
\usepackage{url}            % simple URL typesetting
\usepackage{booktabs}       % professional-quality tables
\usepackage{amsfonts}       % blackboard math symbols
\usepackage{nicefrac}       % compact symbols for 1/2, etc.
\usepackage{microtype}      % microtypography
\usepackage{graphicx}
\usepackage{verbatim} 
\usepackage{algorithm}% http://ctan.org/pkg/algorithms
\usepackage{algpseudocode}% http://ctan.org/pkg/algorithmicx

\algtext*{EndWhile}% Remove "end while" text
\algtext*{EndIf}% Remove "end if" text
\algtext*{EndFor}% Remove "end for" text

\usepackage{subfig}
\usepackage{titlesec}
\usepackage{enumitem}
\usepackage{subfig}
\usepackage{titlesec}
\titlespacing*{\section}{0pt}{2pt}{1pt}
\titlespacing*{\subsection}{0pt}{2pt}{0pt}
\titlespacing*{\subsubsection}{0pt}{0pt}{0pt}
\begin{document}
\bstctlcite{IEEEexample:BSTcontrol}
%\title{\textbf{Invited}- \textbf{NVCell}: Standard Cell Layout in Advanced Technology Nodes with Reinforcement Learning }
\title{\textbf{NVCell}: Standard Cell Layout in Advanced Technology Nodes with Reinforcement Learning }

\author{\IEEEauthorblockN{Haoxing Ren}
\IEEEauthorblockA{\textit{NVIDIA} \\
Austin, TX, USA \\
haoxingr@nvidia.com}
\and
\IEEEauthorblockN{Matthew Fojtik}
\IEEEauthorblockA{\textit{NVIDIA} \\
Durham, NC, USA \\
mfojtik@nvidia.com}
\and
\IEEEauthorblockN{Brucek Khailany }
\IEEEauthorblockA{\textit{NVIDIA} \\
Austin, TX, USA \\
 bkhailany@nvidia.com}
}
\IEEEoverridecommandlockouts
%\IEEEpubid{\makebox[\columnwidth]{978-1-6654-3274-0/21/\$31.00~\textcopyright2021 IEEE \hfill} \hspace{\columnsep}\makebox[\columnwidth]{ }}

\maketitle
%\IEEEpubidadjcol

\begin{abstract}
% with unidirectional metal
High quality standard cell layout automation in advanced technology nodes is still challenging in the industry today because of complex design rules. In this paper we introduce an automatic standard cell layout generator called \textbf{NVCell} that can generate layouts with equal or smaller area for over 90\% of single row cells in an industry standard cell library on an advanced technology node. \textbf{NVCell} leverages reinforcement learning (RL) to fix design rule violations during routing and to generate efficient placements. 

%Automating standard cell layouts is challenging because of the exploding number and complexity of design rule checking (DRC), especially when the design goal is to minimize cell area. %Previous approaches leveraged mathematical optimization methods such as SAT and MILP to find an optimum solution under those constraints. These mathematical optimization methods rely on manual expression of all design rules within an optimization framework and computationally efficient solvers. 
%In this paper we propose a machine learning based approach to handle DRC constrains.  In our approach, we apply a genetic algorithm to create routing candidates and use reinforcement learning (RL) to fix the design rule violations incrementally. A design rule checker provides feedback on violations to the RL agent and the agent learns how to fix them based on the data. This approach is also applicable to future technologies with unseen DRCs. Based on this approach, we built a layout generator called \textbf{NVCell} that includes a device placer based on a simulated annealing method and a router based on a genetic algorithm and reinforcement learning. \textbf{NVCell} can generate layouts with equal or smaller area for over 75\% of cells in an industry standard cell library.

\end{abstract}


%\keywords{Reinforcement Learning, Genetic Algorithm, Standard Cell Routing, 7nm}



\section{Introduction}

%design efforts of standard cell layout: 8-10 months team of 10-15 engineers; need for fast turn around time for circuit optimization; need for early technology development

Standard cells are the building blocks of digital VLSI design. For each technology node, thousands of standard cells need to be designed, which form a standard cell library. Large semiconductor companies and intellectual property providers often have dedicated teams designing standard cell libraries for each technology node. Given a set of design rules, the main cell design objective is minimizing cell width (cell height is fixed for each library) to improve area efficiency. 

Standard cell layout design automation includes two major steps: placement and routing. The placement step places devices and assigns pin locations; the routing step connects device terminals and pins based on net connectivity. Although automated cell layout has been studied extensively, it is still quite challenging on advanced technology nodes. For example the routing step needs to satisfy many demanding design rule checking (DRC) constraints on the lowest metal layers. In advanced technology nodes, not only are the number of design rules exploding, but rules also have become more complex. Most new complexity comes from rules that involve multiple layout shapes that were previously independent from each other. %Although mathematical optimization approaches based on Boolean satisfiability problem (SAT) \cite{SATRyzhenko} and mixed-integer linear program (MILP) \cite{BonnCellCleeff} have been proposed to automate routing and achieved good results, these techniques depend on the assumption that all design rule constraints can be expressed in mathematical forms such as conjunctive normal form for SAT or linear inequality for MILP. It is not clear whether all the DRCs can be expressed efficiently in these forms.  A large number of constraints will have to be created to handle all DRCs, which makes it difficult to scale to larger designs.  Furthermore, these constraints will have to be rewritten by hand for every new technology node or standard cell layout template. 
Another challenge for automating layout is predicting the routability of a generated placement, as not all placements can be routed without violating at least one design rule. It is important to predict routability during placement so that placements can be optimized for routing. To predict routability accurately and efficently is quite difficult.  

Reinforcement learning (RL) has achieved impressive super-human performance in solving many \textit{game} like problems. We are interested in whether RL can also achieve super-human performance in the standard cell layout task. For the routing problem, we believe that the main advantage of RL %over other mathematical optimization methods such as SAT and MILP 
is that it does not require an analytical formulation of the DRCs. The DRC constraints can be enforced by the reward given by a environment where DRC analysis runs independently from the optimization. %\cite{DQNrouteLiao} proposed to apply Reinforcement Learning to the routing problem directly, i.e. make the RL agent create routing actions for each wire where the action space is a routing action (North, South, West, East) for each net. Instead of making the RL agent learn the job of a maze router, which can be solved by many efficient algorithms \cite{VLSISherwani}, 
We decompose the routing problem into two independent steps: initial routing and DRC fixing, and make the RL agent learn to fix DRCs on initial routes. The sub-problem of DRC fixing is easier for RL to learn and it scales to large designs since DRC problems are local. % while general routing requires global information, especially for long routes. 
For the placement problem, we believe RL can help improve runtime by learning a placement policy from data, which was also shown in \cite{RLGoogle}. We also propose to leverage machine learning to predict routability of a standard cell placement. 

Our key contributions are listed below: 
\begin{itemize}[leftmargin=*]
\item We built an automated layout generator, \textbf{NVCell}, which generates competitive layouts for over $92\%$ of cells in an industrial cell library on an advanced technology node. % with area equal or less than the manual design.

\item We propose a simulated annealing based algorithm for device placement and pin assignment. It performs both device pairing and placement concurrently to find a high quality placement. 
\item We propose a machine learning based routability predictor to predict the routability of a given placement. It helps \textbf{NVCell} generate competitive layouts on an additional $9.5\%$ of cells.
\item we propose a RL based placement algorithm to speedup the simulated annealing based placer. It was able to produce the same quality of layout as the simulated annealing based placer on $84\%$ of cells tested after training. 
\item We propose a RL based method to fix DRC errors given existing routes on standard cells. Trained on one standard cell, the model is transferable to all the standard cells we have tested in our library. The model can be further retrained on each cell to improve the results. 
\item We propose a genetic algorithm based routing flow to find minimum routes and optimize the DRC errors. Together with the RL based DRC fixer, it found routable DRC-clean cell layouts with reduced widths compared to the best layouts found by expert layout engineers.

\end{itemize}

\section{Prior Art}

Prior placement techniques include heuristic based methods, exhaustive search based methods, and mathematical programming based methods. The heuristic based methods \cite{chainWimer} first find all possible chains in the circuit, i.e. devices that can share diffusions consecutively, and then select a number of chains that cover all the devices. The exhaustive search based methods \cite{BonnCellCleeff} \cite{NCTUcellLi} go through all possible device placement configurations and might use branch and bound or dynamic programming techniques to speedup the search process. The mathematical programming based methods \cite{MILPAng} \cite{SATJo} leverage MILP or SAT algorithms to find optimal device placement. It has been shown that these placement techniques can generate good placement solutions. One of the key challenges of placement is routability prediction. Some placers use simple heuristics to predict routerbility, some placers actually perform routing to get more accurate estimation. The tradeoff between the routebility predictor's accuracy and runtime should be carefully considered. %However, for our implementation, we prefer a technique that can adapt to custom layout constraints and is also easy to implement. Therefore we propose to use simulated annealing to perform placement.

Prior routing techniques include channel routing, SAT, and MILP based routing methods. Channel routing algorithms have been proposed to do standard cell routing early on \cite{chainWimer}. However, commonly used deterministic channel routing methods such as LEA, Dogleg, Greedy, YACR2, etc. \cite{VLSISherwani}, only generate a particular routing solution and do not handle DRCs well. SAT based routing \cite{SATRyzhenko} creates candidate routes for each terminal pair and leverages SAT to find feasible routing candidates for all terminal pairs. It requires DRC checks to prune all conflicting routing candidates. The quality of routing candidates also limits the final routing quality. Therefore it often can not find routing solutions for complicated cells. MILP based routing methods \cite{BonnCellCleeff}\cite{NCTUcellLi} formulate the routing problem as a mixed integer linear programming problem. It is capable of routing very complex cells in a 7nm process technology. This method, however, relies on a MILP solver to solve a large number of constraints and requires DRCs to be expressed in conditional equality or inequality form. This makes supporting newer technology nodes difficult. \cite{CohesiveJo} uses a combination of MILP and rip-up-reroute techniques to route, which would have similar issues to those mentioned previously.

%Reinforcement learning optimizes an agent's behavior while interacting with its environment. The key elements of RL are action space, observation space, action policy, and reward. At each step, the RL agent can observe the environment from observation space, it then uses its action policy model to select an action in the action space to give to the environment. The environment takes the action and alters its internal state and then gives a reward to the agent for next step. There are many RL algorithms, but most of them can be placed into two categories: value based method such as DQN \cite{DQNMnih} and policy gradient based method such as PPO \cite{PPOSchulman}. Monte carlo tree search (MCTS) \cite{AlphaGoSilver} can also be combined with RL to achieve even better results in unseen conditions.  

RL has achieved great successes in playing games such as Atari and GO. %Recently, researchers also explored using RL to solve combinatorial problems such as TSP \cite{DBLP:journals/corr/BelloPLNB16} and bin packing \cite{binpackLaterre}. 
Recently RL has also been proposed to handle placement and routing problems. \cite{RLGoogle} leverages RL to place macros. \cite{attentionrouteLiao} uses Deep Q Learning (DQN) \cite{DQNMnih} to create routing direction action, i.e. going north, south, etc at each step. \cite{DQNrouteLiao} uses the attention model-based REINFORCE\cite{REINFORCE} algorithm to select routing orders and uses a pattern router to generate actual routes once a routing order is determined. \cite{MCTSrouteHe} leverages both Monte Carlo tree search (MCTS) and neural network based directional action to find routes. Most of these approaches only aim to connect the routes without consideration of design rule violations.  It is not obvious how to extend those methods to handle DRCs for advanced technology nodes. %The application domains of these methods are analog routing or PCB routing, so it is also not clear whether those methods can be extended to the standard cell routing problem. 
Previous machine learning approaches on DRC focus on predicting DRC from early design states, e.g \cite{DRCfloorplan} proposes to predict final DRC from macro floorplan. 
%\subsection{PPO}
%\begin{figure}[]
%  \centering
%  \includegraphics[trim={0.8in 2.9in 1.3in 2.2in},clip, width=0.45\textwidth]{figs/PPO.pdf}
%  \caption{Typical Model Architecture for PPO}\label{fig:ppo}

%\end{figure}

%The key parameters of the PPO models are 
%\begin{equation}\label{eq:PPO}
%    L^{CLIP}(\theta) = \hat{E}_t[min(r_t(\theta)\hat{A}_t,clip(r_t(\theta), 1-\epsilon,1+\epsilon)\hat(A)_t)]
%\end{equation}
%where $\theta$ is the policy parameter which is often the weights of the deep neural network model; $\hat{E}_t$ is the empirical expectation over time;$r_t$ is the ratio of the policy probability under the training and roll-out models, it indicates how different are these two models; $\hat{A}_t$ is the estimate advantage at time $t$, which is a generalized advantage estimation based on equation \ref{}; and $epsilon$ is the hyperparameter, usually 0.1 or 0.2. 

\section{Placement}
Given a set of PMOS and NMOS devices, the goal of placement is to place them on the PMOS row and NMOS row of the cell layout while satisfying technology constraints. In addition to device placement, cell pin locations should also be specified during placement.  

In this section we will introduce two placement methods we have developed for NVCell. One method based on a conventional simulated annealing algorithm and another method based on reinforcement learning. We will also discuss routerabiltiy prediction methods used in placement.

\subsection{Simulated Annealing}
The advantages of the simulated annealing based method are adaptability to custom layout constraints and ease of implementation. Previous placers often separate placement into two steps: pairing and ordering. The pairing step pairs up each PMOS device with a NMOS device to form device pairs. The ordering step generates the placement order of device pairs. The final placement can be inferred from placement order and pairs. These two steps are interdependent, so solving one after the other is sub-optimal. Therefore we design a simulated annealing based algorithm that does both pairing and ordering simultaneously. 

Simulated annealing makes \textit{moves} on a placement representation which specify the placement order of pins, ordering of NMOS and PMOS devices, and whether to flip a device orientation (switching the source and drain positions).  It optimizes a scoring function which is a weighted sum of cell width, routability estimation and technology constraint violations. These \textit{moves} can be categorized either by the types of moves or by the targeted devices of the moves. The \textit{Flip} changes all targeted devices flip flag. The \textit{Swap} swaps targeted devices. The \textit{Move} moves targeted devices to a specific location. The target devices can be either consecutive PMOS devices, consecutive NMOS devices, consecutive PMOS/NMOS device pairs, or pins. 

The simulated annealing algorithm is implemented based on the modified Lam annealing schedule \cite{adaptiveSA} that requires no hyper-parameter tuning. 

\subsection{Routability Estimation} To make the generated placement routable, the key is to include a routability estimation metric in the placement cost function. We first designed a simple routability prediction method. For each net, we draw a horizontal line from its left most terminal to its right most terminal. Then we compute the max and average number of crossing lines on each $x$ position and use a weight average as the routability estimation for the placement. 

This simple routability estimation method works well in our experiments, however, there are still some generated placements can not be routed. Those placements failed routing not because there are not enough wire tracks, but because some terminals can not access the M1 layer. In that case max and average crossing estimation are not adequate to predict the M1 access issues. Therefore we developed a ML based routability prediction model. On each device pair, we collect features such as the number of nets connected to each terminals of the PMOS and NMOS devices, the number of pins near it, and the number of estimated wire crossings over it, etc. We can then concatenate the features of all the device pairs together into a tensor and use 1D convolution and max pooling to predict the routability of the given placement. The routability is labelled as [\textit{routable}, \textit{routable but with DRCs}, \textit{not routable}], which are generated by the router. 

This ML-based routability prediction model is more accurate than the simple prediction method. This model, however, consumes more runtime and makes placement much slower if used in all the steps of the simulated annealing algorithm. Therefore we only use it at the very end steps of the simulated annealing algorithm.

\subsection{Reinforcement Learning}
Simulated annealing works quite well, but for large designs it is relatively slow.  The ML-based routability predictor model makes the runtime even worse. Therefore we also developed a RL-based placer that can generate placement on par with the simulated annealing algorithm with a fraction of the runtime. 
Given a standard cell circuit netlist, the RL placer will place a pair of NMOS and PMOS devices from left to right. The action space of the RL placer is which devices to pair together to be placed at the next step. The action space also includes which pin to be placed next. To support unpaired PMOS or NMOS, we include a dummy PMOS device and a dummy NMOS devices in the the action space, as well as a dummy pin. 

The state space of the RL placer is a graph made of nodes of PMOS, NMOS devices and pins including the dummy devices and pin. The connectivity between the nodes are derived from the net connectivity. For each net, we instantiate two opposite edges between each terminal pair. The node attributes include node types, number of fins for each device, and placement attributes, i.e. whether its placed or flipped. Edge attributes include edge types such as gate-to-source, source-to-drain, etc. Then we leverage the continuous kernel-based convolutional operator \cite{NNConv} to create node embeddings on each node. The convolutional kernels equation is given in Equation \ref{eq1}:
\vspace{-0in}
\begin{equation}\label{eq1}
\mathbf{h}^{\prime}_i = \mathbf{\Theta} \mathbf{h}_i +
\sum_{j \in \mathcal{N}(i)} \mathbf{h}_j \cdot
m_{\mathbf{\Theta}}(\mathbf{e}_{i,j})
\end{equation}
\vspace{-0.15in}

where $\mathbf{h}^{\prime}_i$ and $\mathbf{h}_i$ are the new and previous embeddings of node $i$, respectively; $\mathbf{h}_j$ is the previous embedding of node $i$'s neighbor node $j$; $\mathbf{e}_{i,j}$ is the edge features of edge between $i$ and $j$; $m_\mathbf{\Theta}$ is a linear network that maps edge feature dimensions to node embedding dimensions; and $\mathbf{\Theta}$ is another linear network that maps input node embedding dimensions to output node embedding dimensions. 

\begin{figure}
  \centering
  \includegraphics[trim={0in 1.5in 2.5in 1.5in},clip, width=0.45\textwidth]{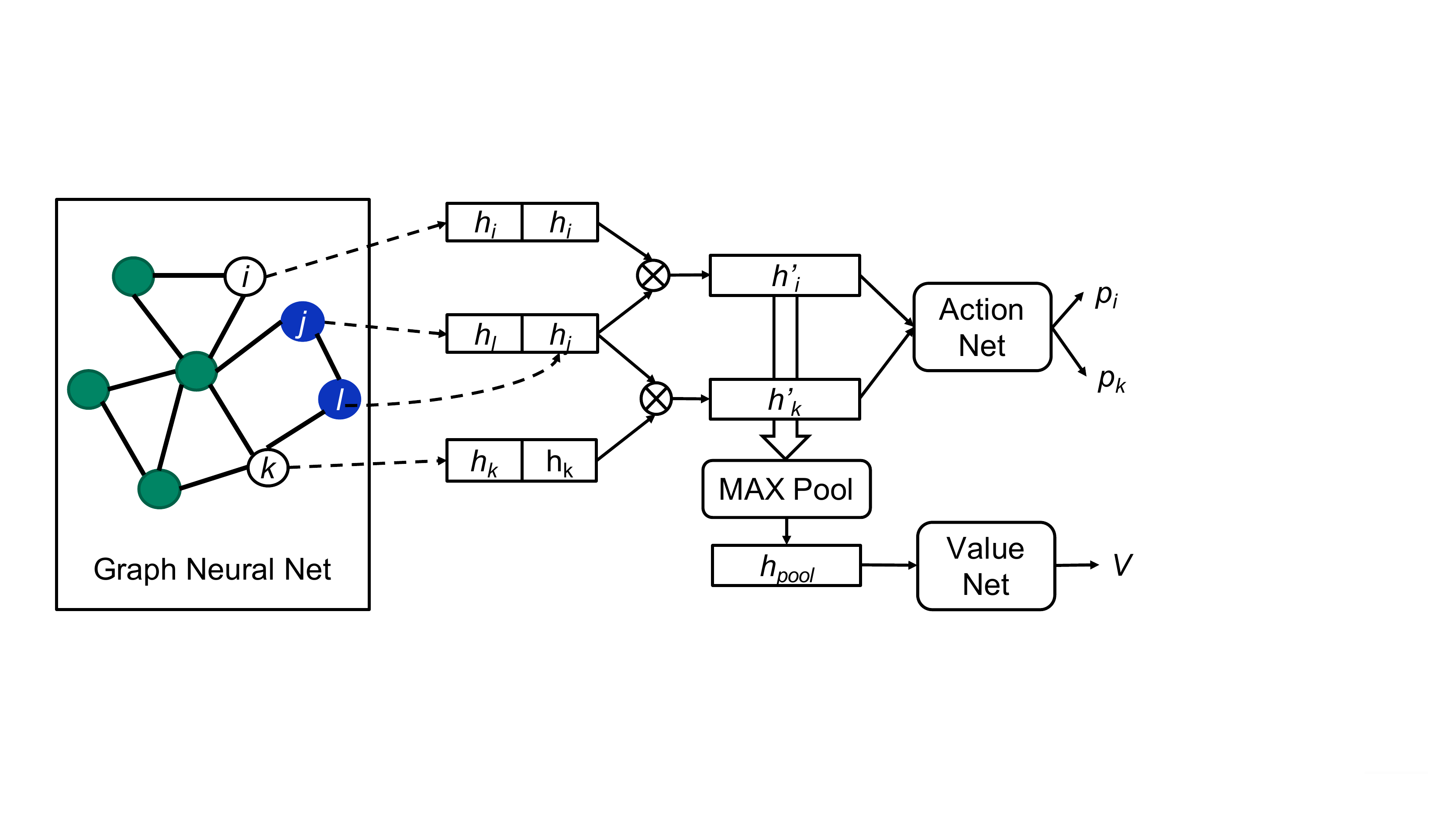}
  \caption{RL Placement Model Architecture. Node $i$ and $k$ are unplaced devices. Green nodes are already placed devices, blue node $j$ and $l$ are the most recently placed PMOS and NMOS device. $h_i$, $h_j$, $h_l$, $h_k$ are the graph embeddings computed by the graph neural network. $p_i$ and $p_j$ are the action probability for $i$ and $j$. $V$ is the value estimation of the current state. Both action net and value nets are two-layer fully connected neural networks. }\label{fig:place}

\end{figure}

Once we have the node embedding of each device and pin node, we then compute the element wise product between the embeddings of unplaced nodes and the most recently placed nodes. These element wise products become the new embeddings of unplaced nodes and then we apply a linear network to predict the action probability of each node selected in the next step. The reason behind this product operation is because the next device pair to be placed should be highly correlated to the most recently placed devices. The model architecture is shown in Figure \ref{fig:place}. 

The reward of the RL environment is derived from the cost function of the simulated annealing method. To minimize the cost, the rewards are all given as negative of the cost function. 

\section{Routing} 
Routing has two steps: a genetic algorithm-based routing step and a RL-based DRC fixing step. The genetic algorithm drives a maze router to create many routing candidates, and the DRC RL agent reduces the number of DRCs of a given routing candidate.

The DRC RL agent only fixes M1 DRC errors. M1 is the lowest routing layer with the most difficult DRC issues. Other DRC errors can be easily pruned out during maze routing. The RL \textit{game} is to incrementally add additional M1 routing segments in order to reduce M1 DRCs. The observation space of the game includes the routes in M1, the DRC positions, and routing mask. The action space is the M1 grid that will be routed next. The rewards include a small negative reward given at each step and a large positive reward associated with DRC reduction.

\begin{figure*}[t]
  \centering
  \includegraphics[trim={0in 1.5in 0in 1.5in},clip, width=\textwidth]{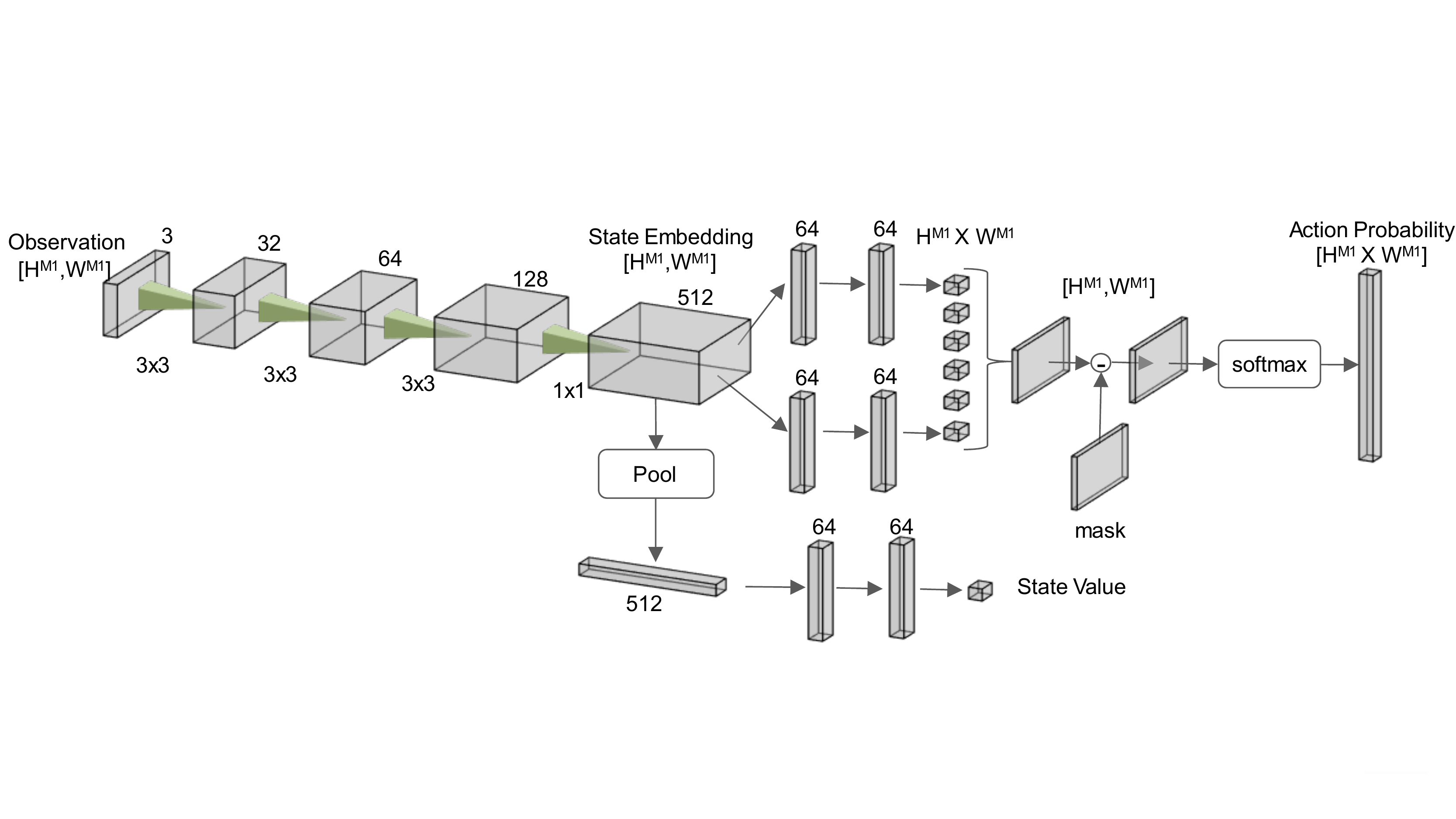}
  \caption{DRC RL Model Architecture}\label{fig:model}

\end{figure*}

We use the Proximal Policy Optimization (PPO) \cite{PPOSchulman} algorithm to implement the RL agent. To build the policy and value networks for PPO, we need to consider two requirements: first it should be invariant to the number of nets, and second it should be invariant to the cell width, i.e. $W^{M1}$. Cell height $H^{M1}$ is a constant for a given library. Our network design is shown in Figure~\ref{fig:model}.

The genetic algorithm based routing algorithm uses routing segments as the genetic representation in that it ensures that good \textit{routing islands} in the routing structure are preserved during genetic operations such as crossover and mutation. The fitness of each individual chromosome in a generation is evaluated based on two metrics: the number of unrouted terminal pairs and the number of DRCs. Other metrics can be also added into fitness function, such as total wiring cost or Design For Manufacturing (DFM) metrics. The complete routing flow is shown in Algorithm \ref{alg:routing}. Routing algorithm and model details are described in \cite{rlrouter}.

\begin{algorithm}[t]
\caption{\textbf{NVCell} Routing Flow}
\label{alg:routing}
\begin{algorithmic}

  \Require nets $\mathbb{N}$, net terminals $\mathbb{T}_n$, generations $G$, population $K$
  %horizontal cut probability $Prob_h$, crossover probability $Prob_c$, mutation probability $Prob_m$,
 
 \Ensure DRC free routing candidates $\mathbb{R}$

 \State create the terminal pair set $\mathbb{P}$

 \State Maze route initial routing candidates $\{\mathbb{R}_1,...,\mathbb{R}_k\}$ for the first generation with random order

 \For {$g \gets 1$ \textbf{to} $G$ }
 \For {$i \gets 1$ \textbf{to} $K$ }
 \State select $\mathbb{R}_{dad}$ and $\mathbb{R}_{mom}$ from $\{\mathbb{R}_1,...,\mathbb{R}_k\}$
 
 \State $\mathbb{R}' \gets crossover(\mathbb{R}_{dad}, \mathbb{R}_{mom}) $
 \State $\mathbb{R'} \gets mutate(\mathbb{R'})$
 \State  $\mathbb{R'}_i \gets$ Maze route unrouted pairs with random order 
 \State run RL DRC fixer on route $\mathbb{R'}_i$
 \State $DRC(\mathbb{R'}_i) \gets $ remaining DRCs
 \If {$DRC(\mathbb{R'}_i)==0$}
 \State return $\mathbb{R'}_i$
 \EndIf
 \EndFor
 \State $\{\mathbb{R}_1,...,\mathbb{R}_k\}$ $\gets$ Top $K$ fitness from $\{\mathbb{R'}_1,...,\mathbb{R'}_k, \mathbb{R}_1,...,\mathbb{R}_k\}$
 \EndFor
 
\end{algorithmic}

\end{algorithm}

\begin{figure*}[t]
 \centering
    \subfloat[Genetic algorithm generated routes with DRC. Dotted red lines show M1 DRC violations.]{
    \includegraphics[trim={0in 3.1in 0in 3.1in},clip, width=0.91\textwidth,height=0.8in]{ 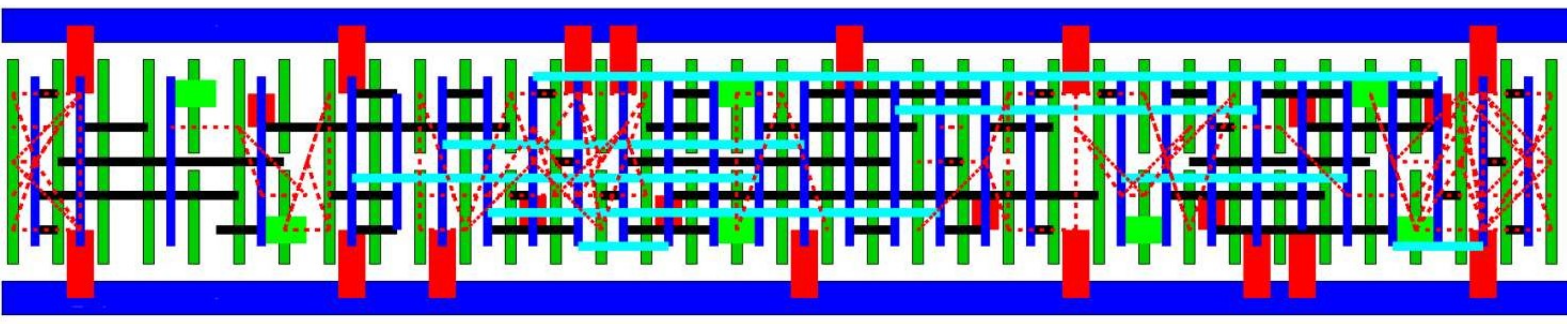}
    }\\[-1pt]
    %\qquad
    \subfloat[Routes after DRC fix by the RL agent.]{
    \includegraphics[trim={0in 3.1in 0in 3.1in},clip, width=0.91\textwidth,height=0.8in]{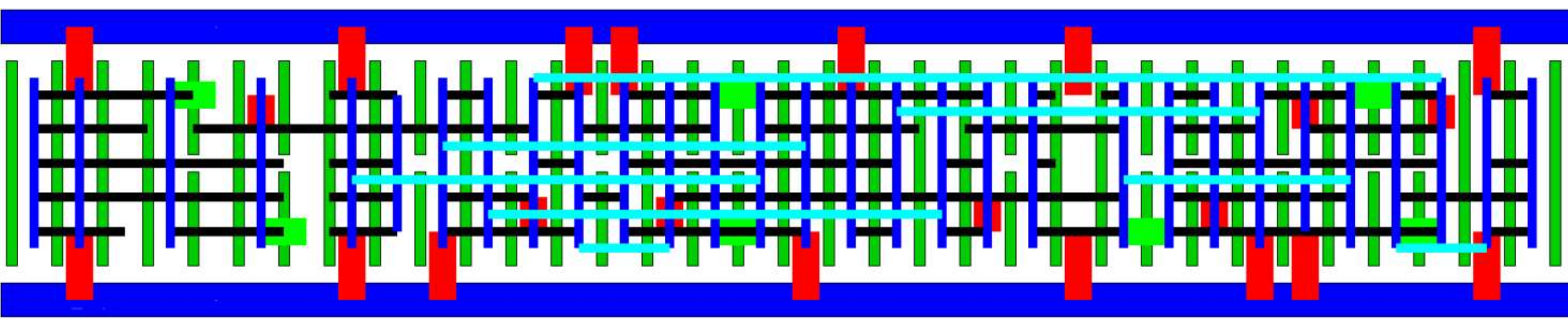}
    }

    \caption{Layouts of a 35-poly wide specialty flip-flop}
    \label{fig:SDFerr}
   
\end{figure*}

\section{Results}

%\begin{figure}[h]
%  \centering
%  \includegraphics[trim={0in 1.2in 0in 1.5in},clip, width=0.45\textwidth]{figs/SDFerr_new.pdf}
%  \caption{Layouts of SDFMEGA. Dotted red lines show DRC violations.}\label{fig:SDFerr}

%\end{figure}
%effectiveness of DRC RL agent, model transferbility
\textbf{NVCell} is implemented with Python and runs on a server with multiple Intel Xeon CPUs where a maximum of 20 threads are used in the implementation. It generates a simplified grid-based cell layout, which is given to a separate Perl program called Sticks to handle DRC checking and conversion to tapeout quality Cadence Virtuoso layout. \textbf{NVCell} produces LVS/DRC clean cells for over 92\% of a total of over 1000 single-row standard cells on an industrial standard cell library (multi-row cells are not currently supported by our Sticks program).  Even more cells can be successfully generated if we relax the width optimization during placement. 15\% of these generated cells have shorter width (between 1 to 8 poly columns) than the same cells in the library, less than $2\%$ of cells have longer width and the rest have the same width. Although not all the cells in the library are designed to have minimum width, we did find meaningful size reduction on some critical arithmetic logic cells and flip-flops. This result shows that \textbf{NVCell} can produce cell layouts competitive with human designers.

For runtime, most of cells run within seconds to a few minutes. The largest cells finish in a couple of hours. We expect to significantly reduce the runtime when the implementations of \textbf{NVCell} and Sticks are switched from Python/Perl to C++. 
 
\textbf{The Routability Predictor} is trained with $90\%$ of random selected cells. For each cell in the training set, we generate 20 unique placements using the simulated annealing algorithm and evaluate the routability of that placement with the router. We route these placement with less routing compute budget than normal to speedup the training data collection and also to make the prediction more pessimistic. We observed test classification accuracy of over $92\%$ on the remaining test circuits. Using the routability predictor allows successful layout to be found for an additional $9.5\%$ of cells.

\textbf{The Placement RL agent} was pretrained with placements generated by the simulated annealing algorithm. The training set consists of $90\%$ of the cells in the cell library. After pre-training, the RL agent is evaluated on the remaining $10\%$ of cells, and it achieved the same optimized width as the simulated anneal algorithm on $84\%$ of cells, and the average width increases only by $1.3\%$. The inference time is in seconds, which is much better than the simulated annealing algorithm which could take more than one hour on complicated cells. The pre-trained RL model can also be retrained on each circuit with PPO2 algorithm. After retraining, the number of cells has the same optimized width as simulated annealing increases to $91\%$ and the average width increase is reduced to $0.6\%$.

 \textbf{The DRC RL agent} was trained only on a flip-flop cell. The environment is initialized with maze routing of all the nets within the cell based on a random net order. Note that the PPO2 algorithm runs multiple environments in parallel, and each environment resets with a new random net order, so the agent will see many different initial routes and different DRC errors. 

Although trained on only one cell, the model is actually transferable to all the cells we have experimented with.  Figure \ref{fig:SDFerr} shows the Stick diagram of both initial routes and routes after DRC fixing for a 35-poly wide specialty flip-flop. The DRC violations, indicated by dotted red lines on Figure \ref{fig:SDFerr}(a), only appear between tracks that are within a certain distance, forming local patterns. These local DRC patterns can therefore be identified by the convolution networks within the RL agent to produce the policy. Because we train the model with many random routes, even only trained on one cell, the agent already sees many different routing configurations, which results in many DRC patterns. These DRC patterns are also invariant between different regions in the same cell or different cells and is why the trained model is transferable to new cells not seen during training.

Both RL agents are implemented with the stable-baselines \cite{stable-baselines} PPO2 (GPU-enabled implementation of PPO) algorithm. The RL environments are implemented based on the OpenAI GYM \cite{openaiBrockman} framework. The training is conducted on a NVIDIA V100 GPU. 

%\begin{figure}[h]
%  \centering
%  \includegraphics[trim={1.5in 1.2in 0.5in 1.5in},clip, width=0.5\textwidth, height=2.2in]{figs/DFQ_new.pdf}
%  \caption{Layouts of DFQ}\label{fig:DFQ}

%\end{figure}
%results of auto placed cells

%discussion about RL: why RL driven routing not working, future work to do RL driven placement

%\begin{acks}

%\end{acks}
\section{Conclusions}
We showed that RL can be leveraged to fix standard cell routing DRCs incrementally from existing routes. We also showed that RL can be used to generate competitive device placements. We have built a layout system called \textbf{NVCell} based on simulated annealing/RL based placement with a ML based routability predictor and a genetic algorithm based router with RL based DRC fixing. We showed that \textbf{NVCell} can generate competitive layouts for a majority of cells in a standard cell library in an advanced technology node. %In the future we plan to research applying RL to the placement of standard cells within a larger block to improve layout quality.   
\small{
\bibliographystyle{IEEEtran}
\bibliography{ref}
}


\end{document}